\documentclass{article}

\usepackage[preprint]{nips_2018}

\usepackage[utf8]{inputenc} 
\usepackage[T1]{fontenc}    
\usepackage{hyperref}       
\usepackage{url}            
\usepackage{booktabs}       
\usepackage{amsfonts}       
\usepackage{nicefrac}       
\usepackage{microtype}      

\usepackage{graphicx}

\title{A neuro-inspired architecture for 
unsupervised continual learning based on 
online clustering and hierarchical predictive coding}

\author{
  Constantine Dovrolis\\
  School of Computer Science\\
  Georgia Institute of Technology\\
  Atlanta, GA 30332 \\
  \texttt{constantine@gatech.edu} \\
}

\begin{document}

\maketitle

\begin{abstract}
We propose that the Continual Learning desiderata can be achieved through a
neuro-inspired architecture, 
grounded on Mountcastle's cortical column hypothesis \citep{Mountcastle:97}. 
The proposed architecture involves a single 
module, called {\em Self-Taught Associative Memory (STAM)}, 
which models the function of a cortical column. 
STAMs are repeated in multi-level hierarchies involving feedforward, lateral and feedback 
connections. STAM networks learn in an unsupervised manner, 
based on a combination of online clustering and hierarchical predictive coding.
This short paper only presents the architecture and its connections with neuroscience. 
A mathematical formulation and experimental results will be presented in an 
extended version of this paper. 
\end{abstract}

\section{Connection with neuroscience}

Instead of providing directly an algorithmic description of STAMs 
(that would hide the connection to neuroscience and cortical columns), 
we first give a sequence of points about cortical columns that 
the design of STAMs is based on.
We should note that the following points are still an active area of research and debate among 
neuroscientists - they should not be viewed as proven facts. 
In the same way that computer science has 
created useful ANNs based on a crude model of a neuron's function, we may also find out that  
STAMs are useful in practice even though they may be only a caricature of how cortical 
columns, and the cortex in general, work.

\begin{figure}
  \centering
   \includegraphics[]{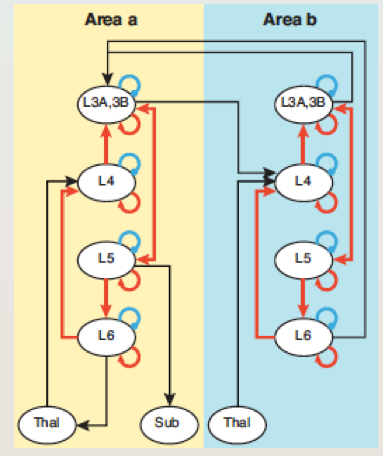}
  \caption{The canonical cortical circuit as represented by Douglas and Martin in
  \citep{Douglas:04}. The diagram shows two instances of the circuit, 
  one in “Area-a” and another in “Area-b”. 
  The layer number (e.g., L4) represents the layer where the soma is located. 
  Red arrows represent excitatory projections between neurons of the same column, while blue 
  arrows represent inhibitory projections (they are mostly between neurons of the same layer). 
  The black arrows represent connections between different columns (or other brain regions).
  Feedforward inputs enter primarily into L4. L4 neurons project to L2/3 neurons. 
  The feedforward outputs originate from L2/3 (pyramidal) neurons. Information from L2/3 neurons 
  is also sent to the deep layers (L5 and L6), where the feedback outputs originate from.
  L6 neurons also project their (intracolumn) feedback back to the input L4 neurons. 
  External feedback connections from other cortical columns project mostly to L3 neurons.}
  \label{fig:corticalcolumn}
\end{figure}

1) The cerebral cortex consists of the same six-layer module, referred to as cortical column, 
repeated throughout the cortex with minor anatomical differences. 
The "canonical cortical circuit" by Douglas and Martin (see Fig.\ref{fig:corticalcolumn}) 
captures what is currently 
known about this module at the level of connections between the six cortical layers 
\citep{Douglas:89,Douglas:04}.
The complete connectome of a cortical column, at the level of 
individual neurons and synapses, is not yet known.  

2) If the same cortical module is used in brain regions associated with very different function 
(e.g., the columns of V1 "see" visual features, the columns of A1 "hear" sounds, 
the columns of the prefontal cortex make plans), 
we are led to the hypothesis that the cortical column performs a very general but 
powerful computational function. 
The neuroscience literature is sparse in offering hypotheses about 
what this common function may be. 
In the following, we refer to this unknown computational function 
of cortical columns as $\Phi(x)$, where $x$ is a vector that represents the collection of 
inputs into a column. 
We propose a specific function $\Phi(x)$ in Section~\ref{sec:stam}.

3) The structure of cortical columns is such that it can be viewed as a module with two input 
channels: a feedforward input channel from lower brain regions (such as the thalamus) or from lower-level cortical regions (e.g., V1 columns projecting to V2 columns), 
and a feedback 
input channel from higher-level cortical regions (e.g., from V2 columns to V1 columns). 
Symmetrically, a cortical column has two output channels: 
the feedback outputs towards lower-level cortical regions and 
other parts of the brain, and the feedforward outputs towards higher cortical regions. These 
feedforward/feedback channels are used to create hierarchies of cortical columns in which most 
connections are reciprocal. 
Based on this distinction, we revise our notation as $\Phi(x_f, x_b)$, where $x_f$ is the 
feedforward input vector and $x_b$ is the feedback input vector. 

4) The internal connectivity of neurons at a cortical column is relatively dense (compared to the 
connection density between different columns) and forms multiple feedback loops. 
In particular, there 
are recurrent circuits of excitatory and inhibitory neurons at layer-4 
(where $x_f$ enters the column), at 
layers 2/3 (where $x_b$ enters the column, and the feedforward output $y_f$ exits the column) 
and at layers 5/6 (where the feedback output $y_b$ exits the column). 
There are also internal feedback circuits from the output neurons 
at layers-5/6 to the input neurons at layer-4. 
Such recurrent circuits and feedback paths from outputs to 
inputs are common in artificial networks implementing sequential/stateful computations, such as 
associative memory networks \citep{Lansner:09}. 
In other words, the highly recurrent structure of cortical 
columns implies that their function is probably more complex than stateless computations 
(such as filtering, feature detection or any other memoryless mathematical transformation of 
their inputs). In fact, 
it has been shown that recurrent neural networks with rational weights are Turing-complete 
\citep{Siegelmann:12}.

5) It has been previously hypothesized, based on the structure of the cortical circuit, that the 
function of cortical columns is to perform predictive coding 
\citep{Bastos:12}. In that 
framework, feedback projections between columns transfer predictions while feedforward projections 
between prediction errors \citep{Rao:99}. A column acts as a generative 
model that can predict its feedforward inputs based on its own priors (stored locally) and 
also based on 
predictions that are fed back from higher-level columns. Note that the predictive coding 
hypothesis 
does not propose a specific algorithm for generating these predictions - it is only a framework 
that 
specifies the type of information (predictions and prediction errors) that flow in the feedback 
and feedforward paths, respectively. 
Our STAM model can be thought of as a specific implementation of the 
predictive coding hypothesis, as described in Section~2. 

6) Cortical columns have the capability to incrementally learn from their inputs, 
storing internal representations that can generalize from few exemplars to useful invariants, 
at least after some initial "development stage". 
For instance, each column of the Inferior Temporal (IT) visual region responds to 
different orientations or partial views of specific animate or inanimate objects (e.g., faces)
\citep{Tanaka:96}.
Each column is highly selective (e.g., it only responding to faces) but it is also has strong 
generalization abilities (e.g., responds to same face independent of rotation, light, occlusions).
In other 
words, it appears that a cortical column stores related "prototypes", and exemplars that are 
similar to that prototype are recognized by that column 
\citep{Kiani:07,Kriegeskorte:08}.
From the computational perspective, this is essentially an {\em online clustering operation}: 
an input vector is 
mapped to its nearest cluster centroid (according to some distance metric). 
Additionally, the centroid of 
the chosen cluster is adjusted incrementally with every new exemplar so that it comes a bit 
closer to 
that input vector - this is how an online clustering module gradually learns the structure of 
the input data. 

7) An online clustering algorithm that is similar to k-means (and asymptotically equivalent to k-
means) can be implemented with a rather simple recurrent neural network of excitatory 
and inhibitory spiking neurons, as shown recently \citep{Pehlevan:18}.
That circuit models the olfactory system in 
{\em Drosophila} but similar recurrent E/I circuits are also present in layer-4 of 
cortical columns.
We hypothesize that the main function of the E/I circuits at layer-4 is also to perform online
clustering of that column's feedforward inputs.

\section{STAM architecture}\label{sec:stam}
Putting the previous seven points together, we now describe the computational 
function $\Phi(x_f,x_b)$ that 
we associate with cortical columns, and describe the proposed STAM module in more detail. 
 

A STAM module receives two input channels (feedforward $x_f$ and feedback $x_b$) 
and it produces the 
corresponding two output channels. 
The first function of a STAM is to perform online clustering of the 
feedforward input vector $x_f$ that it receives. 
We hypothesize that this is also the main function of the E/I 
circuits at layer-4 of a cortical column, 
similar to the neural circuit of \citep{Pehlevan:18}.
The number of clusters in STAMs is dynamically adjusted 
driven by ``novelty detection" 
(if the new input is far from any existing centroid, add a new cluster) 
and ``overlap detection" (if two centroids are quite close, merge the two clusters) - we are 
still investigating how these two mechanisms are implemented in the brain. 

\begin{figure}
  \centering
   \includegraphics[]{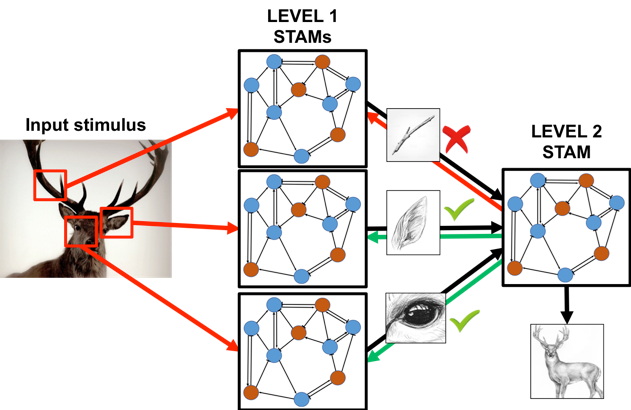}
  \caption{A two-level hierarchy of STAM modules. Each level-1 STAM outputs to the level-2 STAM 
  the closest centroid for the portion of the input covered by its receptive field. 
  The level-2 STAM clusters those centroids and it generates a higher-level prediction of what 
  the entire input image is. It also sends back to each level-1 STAM its prediction for the
  (smaller) receptive field of the corresponding level-1 STAM. The locally generated level-1
  centroid is compared to the predicted centroid from the level-2 STAM, the prediction error 
  is forwarded to higher-level STAMs, while the revised local prediction is forwarded to
  lower-level STAMs.}
  \label{fig:stam-feedback}
\end{figure}

The centroid $c(x_f)$ that is closest to the given exemplar $x_f$ is then compared to the 
predicted centroid that arrives from a higher-level STAM through feedback connections 
(see Fig.\ref{fig:stam-feedback}). 
In cortical columns, this comparison is probably performed at layers 2/3 because that is 
where neurons receive projections from 
both layer-4 and projections from higher-level columns (Fig.\ref{fig:corticalcolumn}). 
In STAMs, this comparison results 
in the difference $c(x_f)- x_b$  between the local centroid and what the next-level 
STAM predicts for the corresponding receptive field. 
It is this difference (prediction error) that constitutes the feedforward 
output of the STAM. 
In cortical columns, we hypothesize that this corresponds to the output projections 
from layers 2/3. 
Returning to STAMs, this prediction error is then transformed to the feedback that will 
be sent back to each of the lower-level STAMs, so that each STAM will only receive the 
feedback that corresponds to its own receptive field. 
In cortical columns, we hypothesize that this is the function of neurons at layers 5/6 
(Fig.\ref{fig:corticalcolumn}).
 
In summary, a STAM module integrates three computational functions: 
online clustering, associative memory formation (i.e., learning and updating the location
of the centroids), and hierarchical predictive coding. 
The online clustering component groups together similar inputs, allowing the STAM to generalize.
The patterns that a STAM 
learns are the centroids of each cluster -- all previously inputs/exemplars are discarded. 
That centroid $c(x)$ becomes the ``recalled memory" when that STAM is presented with vector x. 
Even if the vector $x$ is noisy or partially observed, the centroid $c(x)$
should remain the same as long as $x$ falls in the basin of attraction of $c(x)$. 
Finally, the proposed STAM hierarchies, including the reciprocal projections between 
successive levels, implement hierarchical predictive coding: 
lower-level STAMs reduce the dimensionality of the input data and at the same time they 
are regulated by higher-level STAMs that ``see the bigger picture'' 
(i.e., they have a larger receptive field but potentially in a lower resolution) 
aggregating information from lower-level STAMs.

\section{Connection with Continual Learning}
Let us now examine how the proposed architecture addresses the desiderata
that is often associated with Continual Learning (CL):

{\em 1. Online learning:} STAMs constantly update their centroids with every example.
There is no separate training stage, and there is no specific task for which the network
optimizes the features it learns. 
Any tasks that require classification will of course require one
or few labeled examples so that the corresponding clusters that were formed
previously are now associated with the name of a class.
 
{\em 2. Transfer learning:} The hierarchical nature of the proposed
architecture means that features learned (in an unsupervised manner) at lower-level
STAMs can be reused in different tasks that higher-level STAMs perform.
Through hierarchical predictive coding, this process is also taking place in the
top-down direction: for instance, if the visual data shift at some point 
from bright to dark images, but the objects are still the same (e.g., animals), 
the centroids of the higher-level STAMs will remain the same, modulating the 
lower-level STAMs to darken their centroids instead of learning new prototypes.

{\em 3. Resistance to catastrophic forgetting:} If the system does not operate
at full capacity (see next point), the introduction of a new prototype 
will lead to the creation of new clusters at some STAMs in the hierarchy (e.g.,
layer-1 STAMs will learn new elementary visual features if we start feeding
them natural images instead of MNIST examples -- while a STAM at a higher-level
would create a new cluster when it first starts seeing examples of scooters but
without affecting  the cluster associated with bicycles). 

{\em 4. Bounded system size:} The learning capacity of a STAM architecture depends
on two factors: the number of STAMs and the maximum number of centroids that each STAM 
can store. These two capacity constraints require the system
to forget past prototypes that have not been recently updated with new 
exemplars because the corresponding cluster centroids will gradually shift towards
more recently exemplars of different prototypes. This is a graceful forgetting process however (e.g., gradually forgetting the facial characteristics of our children when 
they were ten years younger).

{\em 5. No direct access to previous experience:} A STAM only needs to store the
centroids of the clusters it has learned so far. Those centroids correspond to 
prototypes, allowing the STAM to generalize. All previously seen exemplars are discarded.

\section{Related work and discussion}
Our main premise is that the cortical column represents the main anatomical and functional 
module in the cortex. 
This premise is inspired by the groundbreaking work of V.Mountcastle 
\citep{Mountcastle:78,Mountcastle:97},
by follow up work by Martin, Douglas, and colleagues
\citep{Douglas:89,Douglas:04},
and by more recent findings such as
\citep{Kaschube:10,Kaas:12,Reid:12,Miller:16}.
It should be noted that this hypothesis is not adopted by everyone in neuroscience - 
there are many ``contrarian voices" that question whether the structure of cortical columns 
is the same throughout the cortex  \citep{Molnar:13}
or whether there is actually a common function behind this structure \citep{Horton:05}.
We believe that this debate reflects the importance of this question in neuroscience. 
We hope to contribute to this debate by exploring the continual learning 
capabilities of hierarchical networks of cortical columns computationally, 
modeling cortical columns as STAMs.

In the context of modeling cortical columns with STAMs, the most relevant prior work has 
appeared in the theoretical neuroscience literature, in the context of hierarchical Bayesian
inference \citep{Lee:03}
and predictive coding \citep{Rao:99,Bastos:12}.
D.Mumford had proposed a similar model of how the cortex works based on 
Grenader's "pattern theory" (without using the term "predictive coding" though) 
\citep{Mumford:92} -- 
that model associates cortical feedback paths with "analysis by synthesis", i.e., 
higher level cortical regions generate hypotheses for the inputs received by lower level regions.
Similar ideas have been proposed by S.Ullman (ascending and descending cortical 
streams performing "pattern search") \citep{Ullman:95},
and by S.Grossberg in a model of ``laminar cortical circuits", which consider the 
connectivity between layers in cortical columns, and bidirectional ``adaptive 
resonance" networks that model the effect of top-down attention mechanisms 
\citep{Grossberg:07}.
More recently, K.Miller has proposed that two operations performed by cortical columns are 
a ``feedforward computation of selectivity" (similar to clustering operations in STAMs) and a
recurrent computation of adaptive gain control between external stimuli and competing 
internally generated signals (similar to the STAM operation of comparing the result of local
clustering with predictions from higher-levels) \citep{Miller:16}.
The idea that cortical columns constitute the building block of associative memory networks, 
which is also central in our STAMs architecture, was first presented by 
Lansner and colleagues \citep{Fransen:98,Lansner:09}.

In the context of machine learning and artificial neural networks, 
the STAM architecture has similarities with several unsupervised methods. 
First, there is a large body of work in clustering-based methods for 
unsupervised learning -- see \citep{Caron:18} for a recent review. 
For instance, the work of Coates and Ng has shown that k-means clustering is not only simpler 
and faster than methods based on sparse autoencoders or Gaussian mixtures -- it also performs
better in feature learning as long as the model has enough hidden nodes (centroids in the
clustering case) and the receptive field (i.e., input dimensionality) 
is sufficiently small \citep{Coates:11}.
For tasks in which the input dimensionality is large, 
clustering can be used as the basic building block of deep hierarchies \citep{Coates:12} --
similar to the feedforward aspect of the STAM architecture. 

The previous clustering-based methods, however, do not include the feedback component of the STAM 
architecture, which is inspired by the recurrent connectivity in the brain. Machine learning
methods such as Helmholtz machines \citep{Dayan:95}
and Deep Predictive Coding \citep{Lotter:16}
are similar to the STAM architecture in terms of how they use feedback connections - but they are 
not based on clustering and they have not been developed in the context of continual learning
tasks, meaning that they assume a mostly stationary environment. 

Another approach to model the function of cortical columns, and to construct hierarchies based 
on that model, has been pursued by J.Hawkins and colleagues at Numenta \citep{George:09}.
Even though our high-level position is the same (namely, that the basic building block of
artificial neural networks should be a functional model of the cortical column, rather than
individual neurons), the Numenta architecture (``Hierarchical Temporal Memory") is significantly
different than the STAMs architecture. One major difference is that the former is based on a
hierarchy of coincidence detectors while the latter combines online clustering and predictive
coding. 

\subsubsection*{Acknowledgments}
This research is supported by DARPA's Lifelong Learning Machines (L2M) program, under Cooperative Agreement HR0011-18-2-0019.

The author is grateful to Sarah Pallas for discussions about cortical columns, and to Zsolt Kira for discussions about the ML aspects of this architecture. Also, thanks are due to Thomas Papastergiou for creating Fig.2 and to Joseph Aribido, Seth Baer, David Nicholson, Astrid Prinz, and James Smith for their comments.  


\end{document}